\documentclass[12pt]{article}
\usepackage{helvet}

\usepackage{natbib}
\usepackage[dvips]{graphicx}
\usepackage[margin=1in,top=0.8in]{geometry}
\usepackage{hhline,amssymb,epsfig,amsmath,array,amsthm,color,setspace,titlesec,lipsum,enumitem,booktabs,subcaption,lscape,float}
\usepackage[hidelinks]{hyperref}
\usepackage[toc]{appendix}
\usepackage[T1]{fontenc}
\usepackage[utf8]{inputenc}
\usepackage{authblk}
\usepackage{changes}
\usepackage[symbol]{footmisc}
\usepackage{fancyhdr}

\usepackage[ruled,vlined]{algorithm2e}
 
\newcommand{\be}{\begin{eqnarray}}
\newcommand{\ee}{\end{eqnarray}}
\newcommand{\ba}{\begin{eqnarray*}}
\newcommand{\ea}{\end{eqnarray*}}

\def\boxit#1{\vbox{\hrule\hbox{\vrule\kern6pt
          \vbox{\kern6pt#1\kern6pt}\kern6pt\vrule}\hrule}}

\def\independenT#1#2{\mathrel{\setbox0\hbox{$#1#2$}%
    \copy0\kern-\wd0\mkern4mu\box0}}

\newcommand{\bOne}{{\bf 1}}

\newcommand{\bp}{{\bf p}}
\newcommand{\bq}{{\bf q}}

\newcommand{\bw}{{\bf w}}
\newcommand{\bx}{{\bf x}}

\newcommand{\bH}{{\bf H}}
\newcommand{\bI}{{\bf I}}
\newcommand{\bP}{{\bf P}}
\newcommand{\bQ}{{\bf Q}}

\newcommand{\bU}{{\bf U}}
\newcommand{\bV}{{\bf V}}
\newcommand{\bW}{{\bf W}}
\newcommand{\bY}{{\bf Y}}

\newcommand {\bfbeta} {\mbox{\boldmath $\beta$}}

\newcommand {\bfSigma} {\mbox{\boldmath $\Sigma$}}
\newcommand {\bftheta} {\mbox{\boldmath $\theta$}}

\newcommand {\bfGamma} {\mbox{\boldmath $\Gamma$}}

\newcommand\iv[1]{{\color{black} #1}} %blue

\newcommand\lc[1]{{\color{black} #1}}
\begin{document}

\title{Improving Sales Forecasting Accuracy: A Tensor Factorization Approach with Demand Awareness}
\singlespacing
\author{}
\author{Xuan Bi\\ \vspace{-4mm} \small Information and Decision Sciences, Carlson School of Management, University of Minnesota\\ Minneapolis, MN 55455 U.S.A.
\and Gediminas Adomavicius\\ \vspace{-4mm} \small Information and Decision Sciences, Carlson School of Management, University of Minnesota\\ Minneapolis, MN 55455 U.S.A. 
\and William Li \\ \vspace{-4mm} \small Shanghai Advanced Institute of Finance, Shanghai Jiao Tong University, Shanghai 200120, China
\and Annie Qu\\ \vspace{-4mm} \small Department of Statistics, University of California, Irvine, CA 92697, U.S.A.}

\date{}
\maketitle 

\clearpage
\section*{Abstract}
Due to accessible big data collections from consumers, products, and stores, advanced sales forecasting capabilities have drawn great attention from many companies especially in the retail business because of its importance in decision making. Improvement of the forecasting accuracy, even by a small percentage, may have a substantial impact on companies' production and financial planning, marketing strategies, inventory controls, supply chain management, and eventually stock prices. Specifically, our research goal is to forecast the sales of each product in each store in the near future. Motivated by tensor factorization methodologies for personalized context-aware recommender systems, we propose a novel approach called the Advanced Temporal Latent-factor Approach to Sales forecasting (ATLAS), which achieves accurate and individualized prediction for sales by building a single tensor-factorization model across multiple stores and products. Our contribution is a combination of: tensor framework (to leverage information across stores and products), a new regularization function (to incorporate demand dynamics), and extrapolation of tensor into future time periods using state-of-the-art statistical (seasonal auto-regressive integrated moving-average models) and machine-learning (recurrent neural networks) models. The advantages of ATLAS are demonstrated on eight product category datasets collected by the Information Resource, Inc., where a total of 165 million weekly sales transactions from more than 1,500 grocery stores over 15,560 products are analyzed.

\noindent\textbf{Keywords:} design science, machine learning, sales forecasting, tensor decomposition, consumer demand

\clearpage
%\title{A Temporal Recommendation Engine for Product Sales Forecasting}
%\maketitle

%\vspace{-4cm}
\section{Introduction}
Supply chain and inventory management involve many complex problems, where decision makers (e.g., producers, distributors, store managers) typically need to consider a wide variety of aspects, including costs, inventory levels, transportation, labor, supply and demand trends, potential risks and gains. Product sales forecasting is a key component in many decision processes, and making accurate sales forecasts constitutes an important and challenging problem.
\lc{For example, store managers often have to decide on the optimal inventory levels of products by making trade-offs between 
	%	product exploitation and exploration. Specifically, the trade-off would be between 
	continuously re-stocking existing items and testing out alternative products to increase novelty, diversity, and serendipity. Effective forecasting techniques are needed, especially when there is little or no direct historical sales data for a specific product in a specific store.}  %For this purpose, we turn to the techniques and methodologies of recommended systems.

One possible solution is to leverage sales data from similar stores and/or from similar products. Store managers could potentially acquire historical data of other stores from market research or consulting companies. Based on the trend and seasonality of product sales in other stores, a store manager could more accurately predict the future sales of a product in her store. Inspired by recommender systems \citep[e.g.,][]{aggarwal2016recommender}, when there are rich data about product sales from other stores, one could consider employing collaborative filtering techniques \citep{breese1998empirical,su2009survey,walter2012moving,lee2019recommender} to take advantage of such information. 
%Specifically, collaborative filtering predicts personal preferences of an individual consumer for items that s/he has not yet experienced, based on the system's knowledge about prior purchases, consumptions, and other interactions of this user, as well as those of other similar users.
%In addition to being valuable to businesses, 
The value of a collaborative recommender system for consumers is that it helps them deal with the information overload \citep{jacoby1984perspectives,horrigan2016information}.
%which comes from the overwhelming number of available choices. 
In other words, it 
%can serve as an information filter for users that 
removes unwanted, irrelevant information and finds matched, personalized products in an efficient manner \citep{chen2005third,liebman2019right}, thus, adding significant value to user experience and increasing product sales \citep{hosanagar2014will}. In the sales forecasting context, similarly, if collaborative filtering is incorporated properly, store managers could use it to filter out unpopular or non-profitable products and hence identify potentially profitable products from a great number of alternative choices.

However, the context of sales forecasting is different from recommender systems, and hence off-the-shelf collaborative filtering methods may not be directly applicable or effective. One major difference is that forecasting targets on predicting \textit{future} values, whereas most preference-based recommendations aim at predicting users' preferences over their unexperienced items, although time can still be taken into account as an important factor for quantifying dynamic user preferences \citep[e.g.,][]{koren2009matrix,koren2010collaborative,sahoo2012hidden,aggarwal2016recommender}. Another major difference is that product sales \iv{are also subject to the dynamics of consumer demand}. For example, if one customer bought a TV from Target, s/he probably would not buy it again from Walmart within a short period of time. In contrast, for preference-based recommendation, for instance, one user could have high preferences for (and hence assign high ratings to) many movies. That is, in addition to the collaborative nature, sales forecasting also entails a competitive nature, where increased sales in one store may satisfy a large portion of local consumer demand, and hence result in reduced sales of similar products in other local stores.

Our work follows the design science paradigm \citep{hevner2004design} in that it designs a data-driven, machine-learning methodology to improve forecasting accuracy for product sales. 
%Follow Dunlavy and precisely defines the problem
The problem setting consists of product sales at the store level, that is, weekly (or daily, monthly, etc.) transactions from $n$ brick-and-mortar grocery stores of $m$ products over weeks $1,2,\ldots,T$. For example, a sample data point could be ``store \#1 sold 80.95 dollars of beer \#246 in week \#3''. In this paper, our goal is to forecast the future sales of every product in every store in weeks $T+1$, $T+2$, etc.
%In other words, we propose a novel model which enables us to forecast the sales in each store for each product (including products which have not been sold in the store before) over different timeframes of interest.

The key contribution of this design artifact is that we incorporate \iv{the element of consumer demand dynamics} into our forecasting model, which allows for a better understanding of the dynamics of the sales competition, as well as provides higher accuracy for predicting future sales.
%This type of predictive capability is useful for different decision makers; for instance, store managers can identify potentially profitable products that have not yet been introduced, and distribution managers can find more suitable stores as outlets for their products in a more objective, data-driven manner. 
This artifact also allows store or distribution managers to forecast the product sales from their competitors, which provides informative implications for planning promotion, logistics, and other business strategies ahead of time.
%Since the product sales data usually exhibit strong time trends and seasonal patterns, one of the key features of the proposed method is the capability to incorporate such critical information in order to make better predictions. 
%Meanwhile, since the proposed method does not intend to provide recommendations for end customers, it is not influenced by unavailable customer information. 
%Meanwhile, a strong prediction accuracy demonstrates that recommender systems are powerful tools for product forecasting.

%Technical introduction of the proposed method
Technically, the proposed method utilizes a regularized tensor (multi-dimensional array) decomposition approach. In our setting, \textit{store}, \textit{product}, and \textit{time} represent three modes of a tensor, and we consider the Candecomp/Parafac (CP) decomposition \citep{carroll1970analysis,harshman1970foundations} to extract \textit{store}-, \textit{product}-, and \textit{time}-specific latent factors. To incorporate \iv{the dynamics of consumer demand}, a novel regularization function is employed which imposes correlation among store-specific latent factors for stores within the same geographical region, as will be discussed later in more detail.
%where negative correlation represents the relatively constant local consumer demand.
After tensor decomposition, we \iv{propose to extrapolate latent factors to future periods via two canonical models} -- seasonal auto-regressive integrated moving-average models \citep[SARIMA; e.g.,][]{wei1994time} or long-short-term-memory recurrent neural networks \citep[LSTM;][]{hochreiter1997long,gers1999learning}. This achieves sales forecasting over the timeframe which is beyond what has been observed in data.
%\textbf{To the best of our knowledge, the proposed method is the first that applies recommender systems to product forecasting, which distinguishes it from most existing methods.}

To illustrate the capabilities of the proposed approach, we use a subset of a rich data collection from the Information Resource, Inc. (``IRI'') \citep{bronnenberg2008database}. The data include 164.9 million store-level weekly transaction records across 47 U.S. markets from 2008 to 2011. In this dataset, more than 1,500 grocery stores are tracked continuously and the sales of 15,560 products are recorded. The products can be classified into 8 categories, including razor blades, coffee, deodorant, diapers, frozen pizza, milk, photography, and toothpaste. Here the eight categories are chosen to span a wide range of consumer packaged goods and have varying product diversity and sample size. In our analysis, we illustrate how to forecast the sales of each product in each store over the next few weeks.
The proposed Advanced Temporal Latent-factor Approach to Sales forecasting (ATLAS) is compared to traditional time-series models as well as recent and competitive latent factor models which demonstrate effectiveness in sales forecasting. The results show that ATLAS is able to substantially improve upon existing methods in terms of sales forecasting accuracy.

%(\nd{***Shall we add more product categories? I've tried 12 categories in total, our method in 8 categories performs well and is reported.***})

%Advantages of the proposed method
The high-level, managerial merit of ATLAS \iv{is in its accuracy of predicting future sales via incorporating local-demand-related information, which can facilitate or improve companies' decision making for production planning, marketing strategies, inventory controls, and supply chain management.}
As one representative example, for many brick-and-mortar stores, the space on shelves and in inventories is usually limited; thus decisions regarding what products to stock is of great importance to many store managers. Through forecasting product sales, ATLAS enhances decision-making on whether, which, when, and where products should be stocked, and tracks the products' potential sales in the upcoming time periods.
%The other is the fast and stable convergence of its algorithm, which is highly valuable for real-time operations. 

%The rest of this article is organized as follows. Section 2 provides an overview of relevant prior works. Section 3 introduces the idea of using \xb{latent factor} methodology for product sales forecasting and presents the proposed approach. In Section 4, we describe the IRI marketing dataset as a context for our data-driven analysis. In Section 5, we discuss the product sales forecasting results of the proposed method and baseline approaches. Section 6 further demonstrates the effectiveness of the proposed method specifically in the context of first-exposure product introduction. Section 7 provides the summary of research contributions, managerial implications, and discussion of \xb{limitations and} future research directions.

\section{Related Literature}
In this section, we review some related literature on product sales forecasting, general machine-learning-based forecasting models, and collaborative filtering techniques which influenced the framework for ATLAS. 

%{\color{red}P.S.: Also cite some Dokyun Lee's paper and Maytal Saar-Tsechansky's paper.}

\subsection{Product Sales Forecasting}

Sales forecasting has been a task of long-standing importance. An accurate sales forecasting conveys important information on investors' future earnings \citep{nichols1979security,penman1980empirical} and can provide managerial implications for companies' inventory management \citep{cui2018operational}, budgeting, marketing, production, and sales planning.
Forecasting models are applied in many stages of a product introduction process \citep{mahajan1988new}, among which forecasting models play an important role in organizations \citep{mentzer2004sales} and are commonly used by managers \citep{mas2014review}. 
%\xb{\cite{kahn2010new} suggests that forecasting models could be categorized as} judgmental approaches, customer/market research techniques, and quantitative methods.

%Several surveys and analyses indicate that judgmental approaches are considered as indispensable for forecasting \citep{lawrence2006judgmental}. Some popular judgmental approaches include panel consensus, historical analogy, grass root forecasting \citep{chase2004operations}, and the Delphi method \citep{brown1968delphi,linstone1975delphi}.
%Furthermore, customer/market research techniques also play an important role in product forecasting. For example, \cite{urban1996premarket} describe the challenge of forecasting customer reaction for a new product and design a new market management system. \cite{fader2003forecasting} examine the impact of marketing variables and the length of a test market on forecasting accuracy, and \cite{amrute1998forecasting} considers promotional events in their forecasting method.

%HERE: cite William suggested OM book

Meanwhile, product sales forecasting heavily relies on quantitative methods \citep{schroeder2013operations}, which typically involve statistical, machine learning, econometric, and optimization models \citep{mas2014review,box2015time}. 
More recently, a number of advanced machine learning methods including neural networks have been developed for sales forecasting \citep{chu2011dynamic,parry2011forecasting,kaneko2016deep}, and
%borrow contextual information to improve forecasting accuracy and 
are applied to both online and in-store sales \citep{walter2012moving}. 
%For example, \cite{chu2011dynamic}, \cite{parry2011forecasting}, and \cite{kaneko2016deep} consider neural networks in forecasting new product sales. 
%\cite{walter2012moving} discuss the feasibility of applying recommender systems to retail stores. 

Furthermore, an emerging direction for improving sales forecasting is through taking advantage of social media information and sentiment analysis \citep{liu2016structured,lau2018parallel}. For example, 
%\cite{liu2016structured} find that the volume of Tweets and the sentiments therein significantly improve forecasting accuracy on online products because of their timeliness. 
\cite{cui2018operational} show that social media information significantly improves accuracy of online retail forecasts. 
\cite{chong2017predicting} and \cite{see2018customer} illustrate the impact of customer reviews on sales forecasting. 
%\cite{lau2018parallel} propose a novel big-data-analytics methodology to mine consumer sentiments for accuracy improvement. 
See \cite{choi2018big} and references therein for a review of other forecasting methods for large-scale sales data.

Existing studies also investigate the association between sales forecasting and its context \citep{Gaur2007estimating,kesavan2010inventory,kremer2011demand}. For example, \cite{osadchiy2013sales} investigate the association of financial market information and retail sales. \cite{curtis2014forecasting} discuss the influence of sales forecast on firms' financial statements. 
%Meanwhile, \cite{Gaur2007estimating} suggest using dispersion among forecasting experts to estimate demand uncertainty. \cite{kesavan2010inventory} incorporate inventory and gross margin in sales forecasts. And \cite{kremer2011demand} investigate forecasters' systematic behavioral biases via a controlled lab experiment. 

In fact, very few existing studies directly incorporate sales competition or consumer demand information in product sales forecasting. 
%One possible reason is that sales forecasting is rarely considered for multiple-store and multiple-products simultaneously beyond tensor and latent factor modeling, while another possible reason is that the quantification of sales competition might be challenging.
\cite{wacker2002sales} discuss the relationship between sales forecasting and resource planning from a managerial perspective.
\cite{sun2008sales} design a neural network to investigate factors that are associated with demand in fashion retailing.
\cite{ma2016demand} study the contribution of stock-keeping-unit-level promotion information to forecasting accuracy via variable selection.
\cite{ferreira2015analytics} and \cite{fisher2018competition} incorporate competition in the retail sales context, but largely focus on the dynamic pricing issues.
And \cite{pavlyshenko2019machine} consider distance to competitor's store as one of the explanatory variables in sales forecasting. 
To the best of our knowledge, however, none of the existing methods aim at forecasting in a multiple-store and multiple-product setting while incorporating sales competition as a key feature to improve accuracy. This is the motivation of our work.

%Recommender systems \citep{aggarwal2016recommender} are known for their highly efficient quantitative methodologies in building accurate predictive models from sparse, limited data.  These techniques have not been used for product sales forecasting; however, we believe that they can offer significant value in this area and, therefore, investigate their applicability in this paper.

\subsection{Machine-Learning-Based Forecasting Models}

In addition to product sales, forecasting as an important and practical goal has been discussed broadly in the statistics and machine learning research areas.

One classical yet canonical forecasting model is the seasonal auto-regressive integrated moving average model \citep[SARIMA,][]{wei1994time}. The SARIMA model is a linear statistical model, which is considered to be general-purpose in the classical time series analysis field. Many traditional forecasting methods are special cases of SARIMA. Given a time series, SARIMA takes account of trend, time lags, auto-regression, moving average, and seasonality, and hence provides effective model fitting and forecasting. It is also computationally efficient and can be easily implemented.

Another model is the long-short-term memory (LSTM), which is a special type of the recurrent neural network. LSTM has demonstrated its effectiveness in many fields, including sequential data analysis \citep{hochreiter1997long,gers1999learning}, handwriting recognition \citep{graves2008novel}, multimodal learning \citep[e.g., image plus text,][]{kiros2014unifying}, speech recognition \citep{sak2014long,li2015constructing}, and anomaly detection \citep{malhotra2015long}. In contrast to the SARIMA model, LSTM is essentially a non-linear machine-learning model. The LSTM model has an internal self-loop to preserve non-zero gradients, and hence partially solves the vanishing or exploding gradients problem commonly seen in recurrent neural networks \citep{rumelhart1988learning}. 
%It has also been mentioned as arguably ``the most commercial Artificial Intelligence achievement'' \citep{vance2018this}. 
Our work considers both SARIMA and LSTM to achieve forecasting.

Recently, additional advances in neural networks have been developed for the forecasting tasks. For example, \cite{shi2015convolutional} propose the Convolutional LSTM to build an end-to-end model for spatio-temporal sequence forecasting. \cite{shi2017deep} propose the Trajectory GPU to formulate location-variant structure in high-resolution forecasts. \cite{wang2019deep} design a deep hybrid model which captures complex patterns and estimates uncertainty simultaneously.

%As mentioned in several of the aforementioned studies, 
Forecasting is of great importance to companies' decision making. Improvement of the forecasting accuracy, even by a small percentage, can have a potentially huge impact.
%on production and financial planning, marketing strategies, inventory controls, supply chain management, customer satisfaction, and eventually companies' stock prices. 
\iv{Our work follows this direction, but focuses on incorporating the element of local consumer demand when historical data from similar (or different) stores and products are available.} The results of our work contribute to this stream of literature by advancing forecasting methodology. % and improving forecasting accuracy.

\subsection{Collaborative Filtering}

%Meanwhile, forecasting, especially multivariate forecasting, can be effectively achieved through machine learning models which are not purposely designed for time series. One of the major tools is CF.
Collaborative filtering (CF) is one of the most popular and effective classes of techniques for personalization as commonly seen in recommender systems \citep[e.g.,][]{ricci2011introduction}. The proposed method utilizes demand-aware tensor factorization, which can be considered as a CF procedure.

%Nevertheless, CF has also been successfully applied to forecasting and shown to be an effective tool \citep[e.g.,][]{koren2010collaborative,dunlavy2011temporal,wang2016recommending,hasan2017collaborative}. 
%In this subsection, we discuss the principles of CF, its advantages, and existing literature. Due to the special background of CF, some of the contexts below are associated with recommender systems.
%The basic principle of CF is to make decisions based on similar users' behavior. For example, if Alice and Bob have similar shopping histories for a number of products that both of them have purchased, then Alice's purchase of certain other products may imply that Bob will purchase them as well. 

A variety of CF-based methods have been developed over the past two decades, for example, nearest-neighbor-based methods \citep{resnick1994grouplens,breese1998empirical,sarwar2001item,bell2007scalable,koren2010factor}, 
%singular value decomposition \citep{funk2006netflix,koren2009matrix}, 
restricted Boltzman machines \citep{salakhutdinov2007restricted}, and
%Several of them have been proved very effective in large-scale, real-world tasks, as was demonstrated by the Netflix Challenge \citep{bennett2007netflix}. 
ensemble methods \citep{jahrer2010combining}.
Moreover, \cite{sahoo2012research} generalize CF to accommodate multi-component ratings. \cite{wang2015collaborative} propose Collaborative Deep Learning to jointly conduct deep learning and collaborative filtering when auxiliary information is available. And \cite{wang2016collaborative} design a Collaborative Recurrent AutoEncoder to jointly predict user ratings and generate content sequences.
%See \cite{cacheda2011comparison}, \cite{feuerverger2012statistical}, and \cite{koren2015advances} for excellent reviews on collaborative filtering techniques.

In particular, singular value decomposition (SVD) 
%-- often referred to as matrix decomposition or factorization -- 
is one of the most widely used CF procedures \citep{funk2006netflix,koren2009matrix,feuerverger2012statistical}. In traditional recommendation applications, SVD formulates user-item interactions in a low-rank utility matrix, and makes predictions through matrix factorization. Some major advantages of SVD include its accuracy and scalability \citep{koren2015advances,clark2016matrix}.
%To achieve this goal, popular regularization methods such as $L_2$-penalty for weight decay \citep{funk2006netflix}, and $L_0$- or $L_1$-penalty for sparsity \citep{chen2001atomic,zhu2016personalized} have been applied. 
Many SVD-related methods have been proposed, such as factorization machines \citep[libFM,][]{rendle2012factorization} and group-specific recommender systems \citep{bi2017group,wang2020logistic}.

\iv{Aside from its methodological advancement, CF has been broadly applied to business settings}. A number of studies investigate its impact on decision-making processes \citep{xiao2007commerce,DBLP:journals/isr/AdomaviciusBCZ13,bi2020consumer}, on sales concentration \citep{fleder2009blockbuster,brynjolfsson2010research,brynjolfsson2011goodbye,oestreicher2012recommendation}, on consumers' willingness to pay \citep{adomavicius2018effects}, and on recommendation diversity \citep{adomavicius2014optimization,muter2017incorporating,song2019diversify}.

Some CF methods also take into account the contextual information \citep{adomavicius2005toward,adomavicius2015context,panniello2016research,bi2018multilayer}.
In particular, time is regarded as a special context of interest in many applications, including ours.
Several approaches have been proposed to accommodate time-awareness \citep[e.g.,][]{koren2010collaborative,koenigstein2011yahoo,sahoo2012hidden,campos2014time,wang2016recommending}.
%In singular value decomposition, \cite{koren2010collaborative} considers time as a continuous variable and utilizes linear extrapolation and kernel methods to adjust it in the baseline fit. In contrast, \cite{koenigstein2011yahoo} categorize time into discrete sessions. \xb{\cite{wang2016recommending} conduct matrix factorization at each time point, and achieve forecasting through vector autoregression.} Also, hidden Markov models can be used to characterize the changes in users' preferences and behavior over time \citep{sahoo2012hidden}.  
In addition, \textit{time} together with \textit{customer}, \textit{product}, and other information can be formulated as a tensor \citep[e.g.,][]{karatzoglou2010multiverse,hidasi2012fast,adomavicius2015context,zhang2020dynamic,bi2020tensors}, which provides for more diverse and flexible interactions. 
%On a related topic, \cite{dunlavy2011temporal} conduct tensor decomposition in social networks for link predictions.

%Nevertheless, CF has also been successfully applied to forecasting and shown to be an effective tool \citep[e.g.,][]{koren2010collaborative,dunlavy2011temporal,wang2016recommending,hasan2017collaborative}. 
CF has also been applied to the forecasting problem. 
For example, \cite{hasan2017collaborative} utilize matrix factorization to formulate weekly energy consumption across multiple households. 
\cite{giering2008retail} achieves retail sales prediction for stores and products for each customer type.
\cite{xiong2010temporal} incorporate time effects via tensor factorization and predict future sales at customer level. And \cite{yu2016temporal} build a flexible autoregressive regularizer and apply matrix factorization to weekly product sales. The proposed ATLAS combine demand-aware tensor factorization with SARIMA and LSTM to achieve forecasting.
%\iv{Aside from the use of tensor factorization, our work considers two canonical, and sophisticated approaches, namely SARIMA and LSTM, to achieve forecasting.}

%Reviewers literature here. 
%Meanwhile, to the best of our knowledge, existing tensor-based, multiple-stores, multiple-products sales forecasting methods do not consider consumer demand information. Some studies have incorporated competition in their retail sales context \citep{ferreira2015analytics,fisher2018competition}; however, they have focused largely on the dynamic pricing issues, and not on complex forecasting problems.

%\subsection{Singular Value Decomposition: A Static Recommender System}
\section{Methodology}%: Introducing Recommender Systems to the Retail Business}
In this section, we first describe one of the fundamental CF-based approaches, namely the singular value decomposition (SVD), and its generalization to tensors.
%We then discuss its interpretation and impact. 
Then we present the proposed method, demonstrate its advantages and effectiveness in predicting future product sales, \lc{and propose two important extensions.}

\subsection{Notation and Preliminaries}

\subsubsection{Singular Value Decomposition} \label{sec:svd}

We first review SVD on a fixed time point. Suppose we have $n$ stores and $m$ products. 
Let an $(n \times m)$-dimensional matrix $\bY_{n \times m}$ denote the utility matrix where each row and each column of $\bY$ represent a store and a product, respectively; and each element $y_{ij}$ represents the (total) sales of product $j$ at store $i$. %The sales could be measured by sales volume, or total dollar sales amount. 

SVD allows $\bY$ \lc{(after standardization, if necessary)} to be factorized into a store-specific latent-factor matrix $\bP$ and a product-specific latent-factor matrix $\bQ$, that is,
$\bY\approx \bP\bQ',$
where $\bP$ and $\bQ$ have a low rank $k \ll \min(n,m)$ \citep{feuerverger2012statistical}.
Specifically, each element $y_{ij}$ is estimated as
\be \label{svd_element}
y_{ij}\approx \sum_{l=1}^k p_{il}q_{jl},
\ee
where $\bp_i=(p_{i1},\ldots,p_{ik})'$ and $\bq_j=(q_{j1},\ldots,q_{jk})'$ are $k$-dimensional latent factors, and are the $i$-th and $j$-th row of $\bP$ and $\bQ$, respectively. We estimate $\bP$ and $\bQ$ such that the distance between $\bY$ and $\bP\bQ'$ is small.
%To implement the SVD, one can apply a stochastic gradient descent algorithm \citep{wu2007collaborative} or simply utilize ordinary least squares in an alternating manner to estimate both $p_i$ and $q_j$ \citep{koren2009matrix}.

It is possible that a certain product is only sold once or twice in a certain store, which could be far fewer than the number of latent factors $k$. Therefore, it is necessary to impose regularization methods to ensure 
algorithm convergence. The simplest regularization method is the $L_2$ penalty (weight decay), as it is convex and has explicit solution for squared loss. That is, we minimize the overall criterion function:
%\textit{***GA: $\Omega$ is used in the formula below, but I did not see it defined/explained before.***}
\be \label{svd-loss}
L(\bP,\bQ)= \sum_{(i,j) \in \Omega}(y_{ij}-\bp_i'\bq_j)^2+\lambda (\sum_{i=1}^{n}\|\bp_i\|^2+\sum_{j=1}^{m} \|\bq_j\|^2),
\ee
where the tuning parameter $\lambda$ is to control the magnitude of the regularization; and $\Omega=\{(i,j): y_{ij} \mbox{ is observed}\}$ is the set of observed sales data. 
An advantageous value of $\lambda$ \lc{(and $k$)} typically can be found automatically using standard practices of machine learning, e.g., by maximizing predictive performance on an independent validation set or through cross-validation. 
%A properly balanced $\lambda$ \lc{(and $k$)} can be selected through cross-validation or minimizing the mean square error on an independent validation set. 
A commonly used alternating least square algorithm \citep[ALS;][]{koren2009matrix}, which estimates $\bp_i$ and $\bq_j$ values iteratively, is described in Algorithm A1 (in Appendix 1) in the supplementary materials.
Then the predicted value for an unobserved element $(i_0,j_0)$ of $\bY$ is given by the estimated $\hat{\bp}_{i_0}$ and $\hat{\bq}_{j_0}$ as:
\be \label{svdr}
\hat{y}_{i_0j_0}=\hat{\bp}_{i_0}'\hat{\bq}_{j_0}.
\ee

\iv{From the business perspective}, $\bp_i$ and $\bq_j$ also describe certain (latent) characteristics of store $i$ and product $j$, respectively. 
In the sales prediction context, $\bp_i$ could hypothetically be interpreted as an indicator of consumer demand from local communities around store $i$. For example, for beverage sales prediction, elements of $\bp_i$ may represent the local demand for the sweetness, coolness, size, or a particular flavor. Of course, elements of $\bp_i$ are latent and solely determined by the algorithm and, thus, may not have a direct mapping into specific features. Nevertheless, if one specific type of ice cream $j$ has its characteristics $\bq_j$ closely match with the demand $\bp_i$, then the SVD model suggests that total sales of ice cream $j$ will be high in store $i$.
We can hence consider the prediction model (\ref{svdr}) as measuring the similarity (unstandardized correlation) between stores and products. 
%In other words, if the highest ``correlation'' between the store characteristics and product characteristics is achieved, then the predicted sales is maximized. %From this point of view, one can draw certain parallels between the SVD approach and traditional content-based recommender systems.

%In summary, we learn the latent factors for individual stores $p_i$'s and individual products $q_j$'s from the observed product sales data, and then are able to make predictions for any unobserved sales (unobserved product-store combinations) based on the learned latent factors.

%\begin{figure}
%	\centering \scalebox{1}{\includegraphics{SVD_process.png}}
%	\caption{\small {\fontfamily{phv}\selectfont Process of the singular value decomposition approach. Store- and product-specific latent factors are obtained from the observed data, and are utilized to predict unobserved sales.}}
%	\label{SVD_process}
%\end{figure}

\subsubsection{Tensor Decomposition} \label{TD}
Now we assume that the sales are time-dependent.
%Each order of a tensor is called a mode.
In addition to \textit{store} and \textit{product}, we assume that the product sales $y_{ijt}$ is also a function of \emph{time} $t$, $t=1,\ldots,T$, where $t$ represents the $t$-th time point, e.g., a calendar week, month, or year, depending on the desired granularity of analysis.
The product sales $y_{ijt}$ can be represented using a third-order tensor $\mathcal{Y}$, where the three modes correspond to \textit{store}, \textit{product}, and \textit{time}.

We consider the CP decomposition \citep{carroll1970analysis,harshman1970foundations}, which generalizes SVD to tensors.
It approximates $\mathcal{Y}$ by $k$ sets of store-, product-, and time-specific latent factors, where, similar to the SVD, $k$ is the number of latent factors. %\nd{See Figure \ref{CPD} in the Appendix for an illustration}.
Element-wise, sales of each product $j$ in each store $i$ at each time $t$ is formulated as
\be \label{svdtr}
y_{ijt}=\sum_{l=1}^{k} p_{il}q_{jl}w_{tl},
\ee
where $\bp_i=(p_{i1},\ldots,p_{ik})'$ and $\bq_j=(q_{j1},\ldots,q_{jk})'$ are latent factors as defined in Section \ref{sec:svd}.  Similarly, $\bw_t=(w_{t1},\ldots,w_{tk})'$ are the latent factors associated with the time effect, e.g., some latent representation of the time trend and seasonality of product sales. Equation (\ref{svdtr}) implies that product sales are based on a ``three-way similarity'', which is generalized from the two-way similarity in the SVD as in (\ref{svd_element}). That is, we expect a high sales volume of product $j$ in store $i$ of time $t$, if the latent characteristics of store $i$, product $j$, and time $t$ are highly consistent with each other. For the same beverage sales example, the demand for the characteristic ``coolness'' might be low during the winter but high during the summer -- such change could be measured by one or more dimensions of the time-dependent latent factor $\bw_t$.
Let $\bW=(\bw_1,\ldots,\bw_T)'$. Then the overall criterion function is given by
\be \label{cpd-loss}
L(\bP,\bQ,\bW)=\sum_{(i,j,t) \in \Omega}(y_{ijt}-\sum_{l=1}^{k} p_{il}q_{jl}w_{tl})^2+\lambda (\sum_{i=1}^{n}\|\bp_i\|^2+\sum_{j=1}^{m} \|\bq_j\|^2+\sum_{t=1}^T \|\bw_t\|^2),
\ee
where $\Omega=\{(i,j,t): y_{ijt} \mbox{ is observed}\}$ is the set of observed data. Similar to the SVD, the minimization of $L(\bP,\bQ,\bW)$ can be achieved through cyclically estimating $\bP$, $\bQ$, and $\bW$, using the blockwise coordinate descent algorithm \citep{friedman2007pathwise}.

\subsection{The Proposed Method: ATLAS}
%In this subsection, \xb{we discuss the proposed ATLAS method, its forecasting, }and its implementation in practice.
%%In contrast to the traditional SVD approach, the proposed method utilizes additional time-dependent information and predicts product sales at future time points.
%%We further assume that the historical data is available from time point $1$ up to time point $T$. 
%Our ultimate goal is to forecast product sales $y_{ijt}$ at $t=T+1, T+2, \ldots, T+h$ for a small integer $h \ge 1$, \xb{regardless of whether store $i$ has sold product $j$ before.}

\subsubsection{\iv{Incorporation of Demand Dynamics}} \label{sec:demand}
%\nd{(***This subsection is mostly rewritten. New narrative sentences are highlighted in blue. Hopefully the readability is improved.***)}

We now discuss the proposed ATLAS approach which incorporates consumer demand. It adopts the same tensor framework as in Section \ref{TD}. That is, store-level data $y_{ijt}$ represents the sales of product $j$ at store $i$ during time $t$. 

\lc{We assume demand dynamics among stores or among products is grouping-specific. Here a grouping can be externally (i.e., based on domain knowledge) defined based on the similarity of attributes. For grouping of stores, such attributes may include geographical regions, store size, reputation, type, or any combinations of them.
	%In our numerical studies, for example, a group is defined as a geographical area with the same 3-digit zip code prefix.
	And for grouping of products, product attributes can be subcategory, volume, price, best-before date, among others. Thus, importantly, the proposed method can incorporate demand dynamics based on groupings of stores, products, or both.}
%In the IRI data analysis, we adopt geographical regions (based on the 3-digit zip code prefix) for store competition groups and subcategories for product competition groups.

%\iv{For \emph{stores},} we typically know which region each store belongs to. The key feature of ATLAS is to incorporate the intuition that each region has its own inherent consumer demand for each product. In our computational experiments, a region is a geographical area defined by the 3-digit zip code prefix. \iv{It should be noted that we do not attempt to model consumer demand directly as in microeconomic studies. Rather, our model borrows the aforementioned intuition to develop a parameterizable, data-driven regularization procedure, which prefers predictive models where the increase (or decrease) of demand for a certain product in one store leads to the decrease (or increase) of demand in other local stores.}

%\iv{For \emph{products}}, we consider data from each product category as a separate dataset, and analyze them individually. In other words, we assume that all products in a dataset belong to the same category, such as \emph{milk} or \emph{coffee}. This is largely for practical purposes, as sufficient transaction data is typically available from each product category for forecasting. However, future studies may incorporate all product categories to leverage cross-category information. %(***\nd{Is this paragraph necessary?}***)

\iv{Specifically, the dynamics of consumer demand is incorporated as follows.} Recall from Section \ref{sec:svd} that $\bp_i$ can be viewed as (latent) consumer demand from local communities around store $i$. Suppose there exist $n_g$ stores in store group $g$ ($g=1,\ldots,G$), namely, $i_1,i_2,\ldots, i_{n_g}$, where each $g$ represents a different geographic region.  
In order to reflect the inherent level of consumer demand in $g$, we propose to
impose correlation among stores' latent factors $\bp_{i_1},\bp_{i_2},\ldots,\bp_{i_{n_g}}$. For example, if such correlation is negative, then increased local demand $\bp_{i_1}$ leads to decreased $\bp_{i_2},\ldots,\bp_{i_{n_g}}$.
%Recall in (\ref{svdtr}) that the product sales $y_{ijt}$ is predicted as a three-way similarity of $\bp_i$, $\bq_j$ and $\bw_t$. For product $j$, therefore, increased $\bp_{i_1}$ leads to increased $y_{i_1jt}$ and decreased $y_{i_2jt},\ldots,y_{i_{m_g}jt}$. 
In other words, if a customer buys a bottle of Heineken in one store, it is less likely that this customer would buy another bottle of Heineken in other stores, which properly reflects the relationship between consumer demand and sales competition. 

Even though basic competition dynamics would typically be reflected by negative correlation, the proposed approach supports \emph{arbitrary} correlation. For example, if we have domain knowledge that store \emph{collaboration} exists among certain pairs of stores (e.g., due to chain-level promotion or complementarity), we may also allow the corresponding correlations to be positive.

\lc{Note that, for products, similarly, we can impose correlation among latent factors $\bq_{j_1},\ldots,\bq_{j_{m_d}}$, corresponding to $m_d$ products in product group $d$, $d=1,\ldots,D$. This allows to formulate product competition or complementality in an analogous manner.}
%\iv{Meanwhile, decreased $\bp_{i_2},\ldots,\bp_{i_{m_g}}$ also lead to decreased $y_{i_2j't},\ldots,y_{i_{m_g}j't}$ for a similar product $j'$. That is, the customer is also less likely to buy other beers in other stores.} %(***\nd{Is this sentence necessary?}***)

%\iv{Next, we illustrate in detail how we impose negative correlation within the tensor framework.} 
Suppose $\bP_g = (\bp_{i_1},\bp_{i_2},\ldots,\bp_{i_{n_g}})$ for a given store group $g$ (e.g., a given geographic region). Let $\bfSigma_g$ be the theoretical covariance matrix of columns of $\bP_g$. For example, the $(1,2)$-th element of $\bfSigma_g$ 
%represents the theoretical covariance between $\bp_{i_1}$ and $\bp_{i_2}$, which 
reflects the perceived demand dynamics between stores $i_1$ and $i_2$. 
Here $\bfSigma_g$ is assumed to be known and positive-semi-definite. Practically, elements of $\bfSigma_g$ would be obtained based on domain knowledge; however, they could also potentially be selected as tuning parameters.
Meanwhile, we let $\hat{\bfSigma}_g$ be the estimated covariance matrix based on the current value of $\bP_g$, that is, $$\hat{\bfSigma}_g = \bP_g' (\bI-\bH) \bP_g,$$
where $\bI$ is a $k$-dimensional identity matrix, $\bH=\bOne(\bOne'\bOne)^{-1}\bOne'$, and $\bOne$ is a $k$-dimensional vector of $1$'s. \iv{In contrast to $\bfSigma_g$, $\hat{\bfSigma}_g$ reflects the actual covariance among $\bp_{i_1},\bp_{i_2},\ldots,\bp_{i_{n_g}}$.}
Our goal is to leverage $\bP_g$ such that $\hat{\bfSigma}_g$ can be as close to $\bfSigma_g$ as possible, especially for the off-diagonal elements across all regions $g$. 
\lc{Similarly, we can define the theoretical and estimated covariance matrix for products as $\bfGamma_d$ and $\hat{\bfGamma}_d= \bQ_d' (\bI-\bH) \bQ_d$ for product group $d$, respectively.}
Then the formulation of tensor decomposition with demand dynamics can be described as
\be \label{dual}
\begin{aligned}
	& \text{minimize}
	& & L(\bP,\bQ,\bW) \\
	& \text{subject to}
	& & \sum_g \|\hat{\bfSigma}_g-\bfSigma_g\| \le c_1, \text{ and } \sum_d \|\hat{\bfGamma}_d-\bfGamma_d\| \le c_2,
\end{aligned}
\ee
where $L$ is defined in (\ref{cpd-loss}), $c_1, c_2 \ge 0$ are constants, and $\|\cdot\|$ is the Frobenius norm (element-wise square loss).
%and each column of $(\bH\bP_g)$ corresponds to the sample averages of each $k$-dimensional latent factor, namely $p_{i_1},p_{i_2},\ldots,p_{i_{m_g}}$, respectively. 

%\ba
%\bR_g=
%\begin{pmatrix}
%1&r_{12}&\cdots&r_{1m_g}\\
%r_{12}&1&\ddots&\vdots\\
%\vdots&\ddots&\ddots&r_{m_g-1,m_g}\\
%r_{1m_g}&\cdots&r_{m_g-1,m_g}&1\\
%\end{pmatrix}
%\ea 
%with $r_{ij} \le 0$ for all off-diagonal elements. 

\iv{However, since demand dynamics exists among groups, most off-diagonal elements in $\bfSigma_g$ (or $\bfGamma_d$) are expected to be non-zero.} Imposing a regularization function to directly shrink the elements of $\hat{\bfSigma}_g$ (or $\hat{\bfGamma}_d$) towards their non-zero counterparts is computationally challenging. For $\hat{\bfSigma}_g$ (or $\hat{\bfGamma}_d$ similarly), recall from linear algebra that $\bfSigma_g=(\bfSigma_g^{1/2})(\bfSigma_g^{1/2})'$ and
\be \label{cholesky}
\|\hat{\bfSigma}_g-\bfSigma_g\| = \Big\|(\bfSigma_g^{1/2})\Big\{(\bfSigma_g^{-1/2})\hat{\bfSigma}_g(\bfSigma_g^{-1/2})'-\bI\Big\}(\bfSigma_g^{1/2})'\Big\|.
\ee
To ease the computational intensity, we revise the problem in (\ref{dual}) based on (\ref{cholesky}) and consider the following problem:
%, given the fact that $(\bfSigma_g^{-1/2})\bfSigma_g(\bfSigma_g^{-1/2})'=\bI$:
\ba
\begin{aligned}
	& \text{minimize}
	& & L(\bP,\bQ,\bW) \\
	& \text{subject to}
	& & \sum_g \|(\bfSigma_g^{-1/2})\hat{\bfSigma}_g(\bfSigma_g^{-1/2})'-\bI\| \le c_{1}, \text{ and } \sum_d \|(\bfGamma_d^{-1/2})\hat{\bfGamma}_d(\bfGamma_d^{-1/2})'-\bI\| \le c_2.
\end{aligned}
\ea 
Therefore, instead of requiring off-diagonal elements of $\hat{\bfSigma}_g$ to be close to those of $\bfSigma_g$ as in (\ref{dual}), now our goal is to shrink off-diagonal elements of $\tilde{\bfSigma}_g \equiv (\bfSigma_g^{-1/2})\hat{\bfSigma}_g(\bfSigma_g^{-1/2})'$ towards off-diagonal elements of $\bI$, which are zero. An analogous goal can be defined for $\tilde{\bfGamma}_d \equiv (\bfGamma_d^{-1/2})\hat{\bfGamma}_d(\bfGamma_d^{-1/2})'$.
%Specifically, 
%where $\bfSigma_g^{-1/2}$ is the inverse of the Cholesky decomposition of $\bfSigma_g$. 
%we consider the transformation $$\tilde{\bP}_g=\bP_g(\bfSigma_g^{-1/2})',$$ and define $\tilde{\bfSigma}_g$ as the sample covariance matrix of $\tilde{\bP}_g$. Now our goal is transformed into shrinking $\tilde{\bfSigma}_g$ to an identity matrix.

Next, we design regularization functions $f_g(\cdot)$ and $h_d(\cdot)$ to achieve this goal. %which shrinks off-diagonal elements of $\tilde{\bfSigma}_g$ towards 0. 
%To simplify the notation, we fix our discussion to an arbitrary yet fixed region $g$ and suppress $g$ from all the subscripts in the rest of this subsection. 
We define $\sigma_{g,ab}$ and $\gamma_{d,ab}$, $a \ne b$, as the $(a,b)$-th element of $\tilde{\bfSigma}_g$ and $\tilde{\bfGamma}_d$, respectively,
and aim at $\sigma_{g,ab} \rightarrow 0$ and $\gamma_{d,ab} \rightarrow 0$.
Then the regularization functions, $f_g$ and $h_d$, which shrink all off-diagonal elements of $\tilde{\bfSigma}_g$ and $\tilde{\bfGamma}_d$ towards 0 can be represented as 
\ba
f_g=\sum_{a \ne b} |\sigma_{g,ab} |, \mbox{ and } h_d=\sum_{a \ne b} |\gamma_{d,ab} |.
\ea
%Notice that the covariance matrix $\tilde{\bfSigma}$ is symmetric and hence $\sigma_{ab}=\sigma_{ba}$.
%Therefore, the regularization function can be re-written as 
%\ba
%f = \sum_{a < b} \|\sigma_{ab}+\sigma_{ba}\| = \sum_{a < b} \| \mbox{vec}(\bP)' \left\{(\gamma_a \gamma_b' + \gamma_b \gamma_a') \otimes (\bI-\bH) \right\} \mbox{vec}(\bP) \|.
%\ea

Through adding $f$ and $h$ into the framework provided by (\ref{cpd-loss}), we have the new criterion function for the proposed method as
\be \label{cpd-loss2}
\begin{aligned} 
	L(\bP,\bQ,\bW)&=\sum_{(i,j,t) \in \Omega}(y_{ijt}-\sum_{l=1}^{k} p_{il}q_{jl}w_{tl})^2+\lambda_1\sum_{g=1}^G f_g+\lambda_1^*\sum_{d=1}^D h_d\\
	&\qquad +\lambda_2 (\sum_{i=1}^{n}\|\bp_i\|^2+\sum_{j=1}^{m} \|\bq_j\|^2+\sum_{t=1}^T \|\bw_t\|^2),
\end{aligned}
%
%	L(\bU,\bV,\bW)&=\sum_{(i,j,t) \in \Omega}(y_{ijt}-\sum_{l=1}^{k} u_{il}v_{jl}w_{tl})^2+\lambda_1\sum_{g=1}^G f_g+\lambda_1^*\sum_{h=1}^H d_h\\
%	&\qquad +\lambda_2 (\sum_{i=1}^{n}\|\bu_i\|^2+\sum_{j=1}^{m} \|\bv_j\|^2+\sum_{t=1}^T \|\bw_t\|^2),
\ee
where $\lambda_2$ controls the magnitude of $\bp_i$, $\bq_j$, and $\bw_t$ values as in (\ref{cpd-loss}), and $\lambda_1$ and $\lambda_1^*$ control the closeness off-diagonal elements of $\tilde{\bfSigma}_g$ and $\tilde{\bfGamma}_d$ to 0, respectively, that is, the degree of demand dynamics.

\subsubsection{Implementation}

In this subsection, we discuss how $(\bP,\bQ,\bW)$ in (\ref{cpd-loss2}) can be estimated.

We adopt a blockwise coordinate descent (BCD) algorithm to minimize the value of $L(\bP,\bQ,\bW)$ in (\ref{cpd-loss2}). The BCD cyclically estimates $\bP$, $\bQ$, and $\bW$.
Since no demand-related regularization is imposed on \emph{time}, the estimation of $\bW$ can be done through ridge regression.
That is, for each time point,
\be \label{svdtw}
\hat{\bw}_{t}=\mathop{\arg\min}_{\bw_{t}} \sum_{\{(i,j):\mbox{ } (i,j,t) \in \Omega\}}(y_{ijt}-\sum_{l=1}^{k} p_{il}q_{jl}w_{tl})^2+\lambda_2 \|\bw_{t}\|^2,\mbox{ }t=1,\ldots,T.
\ee

However, the estimation of $\bP$ or $\bQ$ appears to be more challenging because of the additional penalty imposed by $f$ and $h$. Since the penalty formulates demand dynamics within each group, latent factors of stores or products within each group are no longer independent, and hence the estimation of $\bP$ or $\bQ$ has to be done group by group, instead of store by store (or product by product). For each store group $g$, the estimated $\hat{\bP}_g$ can be derived as
\be \label{svdtp2}
\hat{\bP}_g=\mathop{\arg\min}_{\bP_g} \sum_{i \in \{i_1,\ldots,i_{n_g}\}} \left\{ \sum_{\{(j,t):\mbox{ } (i,j,t) \in \Omega\}}(y_{ijt}-\sum_{l=1}^{k} p_{il}q_{jl}w_{tl})^2 +\lambda_2 \|\bp_i\|^2\right\} +\lambda_1 f_g
\ee
with $f_g$ being represented as
\be \label{f_g}
f_g= \mbox{vec}(\bP_g)'\left(\sum_{a < b} s_{g,ab}\bU_{g,ab}\right) \mbox{vec}(\bP_g).
\ee
Here $\mbox{vec}(\bP_g)$ is the vectorization of $\bP_g$, which stacks the columns of $\bP_g$ into a single vector, and $$\bU_{g,ab}=(\gamma_{g,a} \gamma_{g,b}'+\gamma_{g,b} \gamma_{g,a}') \otimes (\bI-\bH)$$ with $\gamma_{g,a}$ and $\gamma_{g,b}$ being the $a$-th and $b$-th column of $(\bfSigma_g^{-1/2})'$, respectively. And $$s_{g,ab}=\mbox{sign}\Big(\mbox{vec}(\bP_g)' \bU_{g,ab} \mbox{vec}(\bP_g)\Big)$$ represents the sign of each term. 

\lc{For each product group $d$, similarly, we have
	\be \label{svdtq}
	\hat{\bQ}_d=\mathop{\arg\min}_{\bQ_d} \sum_{j \in \{j_1,\ldots,j_{m_d}\}} \left\{\sum_{\{(i,t):\mbox{ } (i,j,t) \in \Omega\}}(y_{ijt}-\sum_{l=1}^{k} p_{il}q_{jl}w_{tl})^2+\lambda_2 \|\bq_j\|^2 \right\} +\lambda_1^* h_d,
	\ee
	where 
	\be \label{h_d}
	h_d = \mbox{vec}(\bQ_d)'\left(\sum_{a < b} r_{d,ab}\bV_{d,ab}\right) \mbox{vec}(\bQ_d),
	\ee
	and $\bV_{d,ab}$ and $r_{d,ab}$ are defined in the same way as $\bU_{g,ab}$ and $s_{g,ab}$, respectively, but with corresponding $\bP_g$ replaced by $\bQ_d$, and $\bfSigma_g$ replaced by $\bfGamma_d$.
	Mathematical derivation of (\ref{f_g}) and (\ref{h_d}) is provided in Appendix 2 of the supplementary materials. Since (\ref{f_g}) and (\ref{h_d}) have a quadratic form, the estimation of each $\bP_g$ and $\bQ_d$ has an explicit solution in each iteration of the BCD algorithm.
	%To solve (\ref{svdtp2}), (\ref{svdtq}) and (\ref{svdtw}), we propose to use the blockwise coordinate descent algorithm.
	%the maximum block improvement algorithm \citep{chen2012maximum}. 
	%We assume that 
	%\singlespacing
	%\ba
	%J_1 &=& 1-L(\hat{P},Q,W)/L(P,Q,W),\\
	%J_2 &=& 1-L(P,\hat{Q},W)/L(P,Q,W),\\
	%J_3 &=& 1-L(P,Q,\hat{W})/L(P,Q,W)
	%\ea
	%%\doublespacing
	%correspond to the relative improvement of the criterion function $L$ via updating $P$, $Q$, and $W$, respectively, and that $J=\max\{J_1,J_2,J_3\}$. At each iteration, only the block with the largest improvement $J$ is updated.
	Then the estimation of $(\bP,\bQ,\bW)$ can be achieved through estimating (\ref{svdtp2}), (\ref{svdtq}), and (\ref{svdtw}) cyclically. The high-level summary of the overall algorithm is provided in Section \ref{sec:fcst} (Algorithm \ref{alg:bcd}).}

%may lead to non-convergence issue or convergence to a point where the criterion function only ceases to decrease \citep{chen2012maximum}. For this reason, we consider the maximum block improvement algorithm here, which guarantees convergence to a stationary point (i.e., to a local minimum along each block) and also exhibits fast convergence, as will be demonstrated in the discussion of empirical results.

\subsection{Forecasting} \label{sec:fcst}

\begin{figure}
	\centering \scalebox{0.5}{\includegraphics{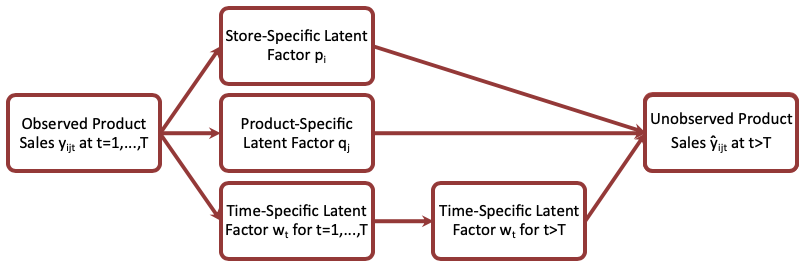}}
	\caption{\footnotesize{High-level process of the proposed method (ATLAS). Store-, product-, and time-specific latent factors are obtained from the tensor decomposition. A time-series model is applied to extrapolate time-specific factors to future time points ($t>T$). The new time-specific latent factors, together with the original store- and product-specific latent factors, are utilized to predict future sales.}}
	\label{TREF_process}
\end{figure}

The previous subsection describes the procedure of how a sparse tensor of sales can be decomposed into store-, product-, and time-specific latent factors, while also incorporating the element of local demand dynamics within each region. 
Suppose our data are collected up to time $T$. The forecasting of future events, say $y_{ij,T+1}$, is not feasible within the tensor directly, since tensor decomposition is not able to estimate latent factors $\bw_t$ at $t>T$. To achieve this, one has to consider a time series model. The entire high-level process is illustrated in Figure \ref{TREF_process}.

One major advantage of tensor decomposition is that we convert a problem of analyzing millions of time series (e.g., combinations of thousands of stores and thousands of products) to a problem of analyzing a $k$-dimensional time series $\bw_t$, where usually $k \le 100$. Some existing approaches can be applied to extrapolate $\bw_t$ from $t \le T$ to $t>T$, for example, the kernel methods \citep{koren2010collaborative}, Holt-Winters method \citep{dunlavy2011temporal}, vector autoregression \citep{wang2016recommending}, or context assertion \citep{hasan2017collaborative}. \lc{Second, sales data contain a large percentage of ``missing values'' where stores may have little or no records of selling certain products. Using time-specific latent factors can borrow data information from similar stores or products, and hence reduce the impact of missing data. Both of these advantages are also noted in \cite{yu2016temporal}.}

\iv{To demonstrate this point, we aggregated store- and product-level IRI data into region- and category-level, respectively, such that sales competition across regions or categories is light and that sales for all region-category-week combinations are observed. Then we find that applying tensor decomposition prior to a time series model is 3\% more accurate and 49\% faster than applying time series models to each combination individually.}

\lc{In this article, we consider two canonical and comprehensive time series models for forecasting. One is the 
	%seasonal auto-regressive integrated moving average (
	SARIMA 
	%or seasonal ARIMA) 
	model \citep[e.g.;][]{wei1994time}. 
	%The SARIMA model is a linear statistical model, which is considered to be general-purpose in the classical time series analysis field. Many traditional forecasting methods are special cases of SARIMA. Given a time series, SARIMA takes account of trend (via differencing), time lags, auto-regression, moving average, and seasonality, and hence provides effective model fitting and forecasting. It is also computationally efficient and can be easily implemented. % via standard software packages. 
	The other is the LSTM neural network \citep{hochreiter1997long,gers1999learning}.
	%The other approach uses the long-short-term memory (LSTM) network, which transforms traditional time series analysis into a supervised learning problem, and is a special type of the recurrent neural network. LSTM has demonstrated its effectiveness as a deep learning method in many fields, including sequential data analysis \citep{hochreiter1997long,gers1999learning}, handwriting recognition \citep{graves2008novel}, multimodal learning \citep[e.g., image plus text,][]{kiros2014unifying}, speech recognition \citep{sak2014long,li2015constructing}, and anomaly detection \citep{malhotra2015long}. In contrast to the SARIMA model, LSTM is essentially a non-linear machine-learning model. The LSTM model has internal self-loop to preserve non-zero gradients, and hence partially solves the vanishing or exploding gradients problem commonly seen in recurrent neural networks \citep{rumelhart1988learning}. It has also been mentioned as arguably ``the most commercial Artificial Intelligence achievement'' \citep{vance2018this}.
	Technical details on how SARIMA and LSTM work are provided in Appendices 3 and 4 of the supplementary materials, respectively.}

For an arbitrary time series $\{x_t\}_{t=1}^T$, both SARIMA and LSTM take $x_1,\ldots,x_T$ as the input, and output $x_{T+1},\ldots,x_{T+\Delta}$ for a small number $\Delta >0$. For the proposed tensor decomposition method, we apply SARIMA or LSTM to each time series $\{w_{tl}\}$, $l=1,\ldots,k$, and acquire its forecasting value at $t>T$.
Then the sales at $t=T+1,\ldots,T+\Delta$ can be predicted as
\be \label{forecasting}
\hat{y}_{ijt}=\sum_{l=1}^{k} \hat{p}_{il}\hat{q}_{jl}\hat{w}_{tl}.
\ee

In summary, ATLAS is achieved through two steps. The first step utilizes tensor decomposition to provide store-, product-, and time-specific latent factors. In the second step, while maintaining the store- and product-specific latent factors unchanged, we extrapolate time-specific latent factors to future time points. Then we forecast future sales through using all latent factors via (\ref{forecasting}). 
Algorithm \ref{alg:bcd} summarizes the entire ATLAS method. 
For the $u$-th iteration, the improvement of the criterion function $J$ is defined as $J=1-L(\bP^{(u)},\bQ^{(u)},\bW^{(u)})/L(\bP^{(u-1)},\bQ^{(u-1)},\bW^{(u-1)})$, where $L(\bP,\bQ,\bW)$ is defined in (\ref{cpd-loss2}).

%The algorithm holds for whatever model is utilized to estimate the time effect.

%{\color{red}May move the following paragraph to somewhere earlier (Sec 3.1?), or Sec 5? or remove?.}
%
%In contrast to treating the product sales of each product at each store as an independent time series $\{y_{ijt}\}_{t=1}^T$, the ATLAS brings three significant advantages. First, the estimation of latent factor vector $w_t$'s incorporates information across \textit{all} available stores and products.\footnote[3]{The same benefit applies to the estimation of the latent factor vectors $p_i$ and $q_j$ -- this is a key advantage of SVD-type methods.} In contrast, each independent time series $\{y_{ijt}\}_{t=1}^T$ for particular product $j$ at particular store $i$ could have many missing observations, which could lead to significant forecasting bias. Second, since $w_t$ is a $k$-dimensional latent factor, the ATLAS method requires applying $k$ seasonal ARIMA models \xb{or an LSTM neural network with $k$ features. In contrast, analyzing each $\{y_{ijt}\}_{t=1}^T$ independently requires calculating $n \times m$ time series,} which is impractical for many real-world applications with large numbers of stores and products. Third, after the tensor decomposition, the time series $w_t$ contains only time-dependent information. In contrast, \xb{a time series model} on $\{y_{ijt}\}_{t=1}^T$ could be confounded with store- or product-specific factors.   All of these advantages contribute significantly towards more accurate and robust sales forecasting.

\vspace{2mm}
%\singlespacing
\begin{algorithm}[H] \label{alg:bcd}
	\SetAlgoLined
	\KwData{$\lambda_1$, $\lambda_1^*$, $\lambda_2$, $k$, $\Delta$, and initial values of $J$, $\hat{p}_i$, $\hat{q}_j$, and $\hat{w}_t$, $t=1,\ldots,T$}
%	\vspace{-2mm}
	\KwResult{$\hat{y}_{ijt}$ for $i=1,\ldots,n$, $j=1,\ldots,m$ and $t=T+1,\ldots,T+\Delta$}
%	\vspace{-2mm}
	%	initialization\;
	\While{$J>1 \times 10^{-3}$}{
%		\vspace{-2mm}
		1: Estimate $\hat{\bP}_g$ for each store group $g$ through (\ref{svdtp2}),\\% and calculate $J_1$\\
%		\vspace{-2mm}
		2: Estimate $\hat{\bQ}_d$ for each product group $d$ through (\ref{svdtq}),\\% and calculate $J_2$\\
%		\vspace{-2mm}
		3: Estimate $\hat{\bw}_t$, $t=1,\ldots,T$, through (\ref{svdtw}),\\% and calculate $J_3$\\
	}
%	\vspace{-2mm}
	4: Apply SARIMA or LSTM to each time series $\{\hat{w}_{tl}\}_{t=1}^T$, $l=1,\ldots,k$\\
%	\vspace{-2mm}
	5. Predict $\hat{\bw}_{t}$ at $t=T+1,\ldots,T+\Delta$\\
%	\vspace{-2mm}
	6: Calculate each $\hat{y}_{ijt}$ through (\ref{forecasting})\\
	\caption{ATLAS with Forecasting}
\end{algorithm}

%\doublespacing

\subsection{Model Extensions}

\lc{\subsubsection{End-to-End Learning}
	
	%\xb{Cite Referee 1's paper here.}
	
	One important extension of the proposed method is its ability to be formulated as an end-to-end predictive modeling technique \citep[e.g.,][]{wang2015collaborative,wang2016collaborative}. As illustrated in Algorithm \ref{alg:bcd}, the current version of ATLAS achieves tensor decomposition and time series analysis in two steps. An end-to-end version of ATLAS allows simultaneous achievement of these two steps. This makes the implementation of ATLAS more efficient, and, practically, may impose fewer technicalities for business analysts.% in real-world data analysis.
	
	Specifically, instead of minimizing a criterion function $L(\bP,\bQ,\bW)$ as in \eqref{cpd-loss2}, we are now minimizing a new criterion function $L(\bP,\bQ,\bW,\bftheta)$ described as follows
	\ba
	\begin{aligned}
		L(\bP,\bQ,\bW,\bftheta)=& \sum_{(i,j,t) \in \Omega}(y_{ijt}-\sum_{l=1}^{k} p_{il}q_{jl}w_{tl})^2\\
		&+\lambda_1\sum_{g=1}^G f_g+\lambda_1^*\sum_{d=1}^D h_d+\lambda_2 (\sum_{i=1}^{n}\|\bp_i\|^2+\sum_{j=1}^{m} \|\bq_j\|^2+\sum_{t=1}^T \|\bw_t\|^2)\\
		&+\lambda_3 \sum_{t=\Delta_0+1}^T\|\bw_t-\xi(\bw_{t-1},\ldots,\bw_{t-\Delta_0}|\bftheta)\|^2,
	\end{aligned}
	\ea
	where $\xi(\cdot|\bftheta)$ is the forecasting model approximated by either SARIMA or LSTM, $\bftheta$ is a vector of all parameters introduced during the training of $\xi(\cdot|\bftheta)$ (e.g., auto-regressive and moving-average coefficients in SARIMA, or weight matrices and bias vectors in LSTM), $\Delta_0 > 0$ is the prespecified time lag, and $\lambda_3 >0$ is a new tuning parameter.
	
	We still consider the blockwise coordinate descent algorithm when optimizing the criterion function above. In other words, parameters $(\bP,\bQ,\bW,\bftheta)$ are estimated cyclically, where the estimation of $\bP,\bQ,\bW$ follows (\ref{svdtp2}), (\ref{svdtq}), and (\ref{svdtw}), respectively, and the estimation of $\bftheta$ is provided by either SARIMA or LSTM. \lc{Numerical results of this end-to-end version of ATLAS are provided in Appendix 5 of the supplementary materials. Importantly, the original and end-to-end versions of ATLAS provide highly similar forecasting performance.}
	
	\subsubsection{Incorporating Contextual Information} \label{contextual}
	
	Another important extension is to incorporate contextual information that could have a direct impact on sales; our proposed approach is able to incorporate this aspect naturally, as part of the pre-model-building preparation.
	%, such as unit price, store or product promotions, among others. 
	In particular, suppose we observe a vector of independent variables $\bx_{ijt}$. For example, elements of $\bx_{ijt}$ may represent the price, and promotion, managerial, and operational strategies of product $j$ in store $i$ at time $t$. Then we can fit a linear regression to control the effects of variables in $\bx$ before applying ATLAS. In other words, we define 
	$e_{ijt} = y_{ijt} -\bx_{ijt} \hat{\bfbeta},$
	where $\hat{\bfbeta}$ is the estimated regression coefficient of $y_{ijt}$'s fitted against $\bx_{ijt}$'s. Next, we replace each $y_{ijt}$ by $e_{ijt}$, and conduct Algorithm \ref{alg:bcd} to forecast $\hat{e}_{ijt}$'s for $t=T+1,\ldots,T+\Delta$. Then the final context-aware forecast can be obtained as
	$$\hat{y}_{ijt}=\bx_{ijt} \hat{\bfbeta} + \hat{e}_{ijt},\mbox{ for }t=T+1,\ldots,T+\Delta.$$
	
	Incorporating contextual information brings two advantages. First, from a managerial perspective, through fitting a regression model, managers could know which factors are significant contributors to sales, as well as how these factors are influencing the sales amount. This allows managers to design more informed pricing and promotion strategies. Second, from a technical perspective, incorporating contextual information may improve forecasting accuracy. Along this direction, we could also consider deep learning techniques such as feedforward neural networks or embedding approaches to further enhance the forecasting results. \lc{Numerical experiments on IRI data incorporating price and promotion information are conducted and provided in Appendix 6 of the supplementary materials.}}

\section{A Real-World Application: IRI Marketing Data}

\subsection{Data Description} \label{sec:data}
\begin{figure}
	\centering \scalebox{0.65}{\includegraphics{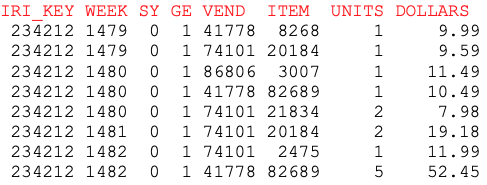}}
	\caption{\footnotesize A snapshot of the IRI data.}
	\label{snapshot}
\end{figure}

In our study, we use the IRI marketing data as an example to demonstrate the effectiveness of the proposed method. The data we acquired are from 2008 to 2011 at a weekly level of granularity (i.e., 208 weeks in total).
Specifically, the dataset contains weekly transactions of 2,447 grocery stores from over 47 U.S. markets. 
%During the 11-year period, 161,114 types of products were sold. 
A detailed description of an early version (2001-2005) of the data, \lc{as well as the data's availability,} can be obtained from \cite{bronnenberg2008database}. Figure \ref{snapshot} illustrates a snapshot of the original data. Here the first column is the de-identified store ID. Notice that, although the data are collected at the store level, all stores in the datasets are chain (rather than independent) stores. \cite{bronnenberg2008database} provide the rationale that independent stores ``are less important for competition in most markets.'' This also aligns with our research goal where important competitors are included in the forecasting. Zipcode of each store is also provided in a separate spreadsheet. The second column represents the week ID, where 1479 corresponds to the first week of 2008 (i.e., December 31, 2007 to January 6, 2008) and 1686 corresponds to the second-to-last week of 2011 (i.e., December 19, 2011 to December 25, 2011, the last week that our data are collected). Columns 3-6 (i.e., system code, product generation, vendor ID, item ID) together make up a twelve-digit Universal Product Code (UPC) which is unique for each product and thus is used as the product ID in our analysis. In the last two columns, the volume and dollar amount of sales of each product are recorded. For example, in the first row in Figure \ref{snapshot}, store \#234212 has sold 1 unit of product \#0-1-41778-08268 in week \#1479 for 9.99 dollars.

Due to the fact that many products come in different sizes but may share the same unit (for example, beverages may come in one single bottle or a six-pack, selling either one of which would be counted as a 1-unit sale, but essentially they are very different), we therefore choose dollar amount as the response variable for analysis and forecasting. Nevertheless, we are aware of the importance of sales units in the sales forecasting area. The proposed method, as well as all the competing methods, can be applied to the sales unit forecasting if they are tuned accordingly.
%\footnote{\xb{At the request of IRI, data from the most recent 2 years are usually not released.}}

%To illustrate the proposed method, we analyze 
Data from eight categories of products are analyzed, including razor blades, coffee, deodorant, diapers, frozen pizza, milk, photography related products, and toothpaste. That is, a spreadsheet similar to Figure \ref{snapshot} is collected for each product category. The proposed method, as well as its competing methods, are applied to each product category individually and compared. In other words, the eight categories are treated as eight independent datasets for the forecasting purposes. %Here the eight categories span a wide range of consumer packaged goods and have varying product diversity (and sample size). This demonstrates the broad applicability of ATLAS across different consumer goods sectors and data scales.

In Table \ref{descriptive}, we list the number of grocery stores, number of products, and sample size (number of rows as in Figure \ref{snapshot}) within each category. 
%For data-driven forecasting purposes, 
For every category, we select stores or products only if they have more than 1,000 or 200 transactions, respectively. That is, on an average week, a selected store sells at least 5 products, and a selected product is sold in at least one store across the entire nation. \lc{We also demonstrate in Appendix 7 of the supplementary materials that the performance of ATLAS does not change substantially if all stores and products are included.}

The number of stores is similar across all datasets except for the photography category. 
%This indicates that most stores have sold products of seven other categories. 
In terms of the number of products, nearly 5,000 coffee products and 4,000 milk products nationwide pass the aforementioned screening. The sample size of each category ranges from 0.2 million to nearly 39 million. However, although the sample size for most datasets is large, there are still a huge number of store-product-week combinations that do not have any sales. For example, there are 1,528 stores that sell milk, 3,739 types of milk, and a total of 208 weeks. The total number of store-product-week combinations could go up to $1528 \times 3739 \times 208 \approx 1.2\mbox{ billion}$. In contrast, the sample size of 26.4 million observed milk-related combinations, although large, is still highly sparse in the store-product-week tensor, as it only consists of roughly $2.2\%$ of the total elements. 

\begin{table}
	\begin{footnotesize}
		\caption{The number of grocery stores, number of products, and the sample size (number of rows as in Figure \ref{snapshot}) are provided for each dataset we analyze.}
		\begin{center}
			
			\begin{tabular}{cccc}
				\hline
				\hline
				&Num. of Stores	&Num. of Products	&Sample Size	\\%&Training Size 	&Testing Size	\\
				\hline
				Blades		&1,417		&686 				&9,001,597	\\%&8,695,431	&306,166			\\
				Coffee	 	&1,535 		&4,844 				&38,971,593	\\%&37,390,382	&1,581,211		\\
				Deodorant	 	&1,502 		&1,445 				&25,869,335	\\%&25,017,729		&851,606			\\
				Diapers		&1,477		&1,653				&13,622,073	\\%&13,177,698		&444,375			\\
				Frozen Pizza	&1,528		&2,057				&28,574,874	\\%&27,519,609		&1,055,265		\\
				Milk			&1,528 		&3,739				&26,439,459	\\%&25,307,676		&1,131,783		\\
				Photography	&176			&131					&241,851		\\%&237,873			&3,978		\\
				Tooth Paste	&1,511		&1,005				&22,173,211	\\%&21,394,782		&778,429		\\
				%total sample size: 164,893,993
				\hline
				\hline
			\end{tabular}
			
		\end{center}
		\label{descriptive}
		
	\end{footnotesize}
	
\end{table}

%%\xb{(This paragraph and Table 2 may be moved to Supp.)} 
%In Table \ref{summary}, we provide the summary statistics regarding the average sales amount of each single product (per store per week) for each dataset. It can be seen that the average sales amount ranges from 6.88 dollars (for the deodorant category) to 110.14 dollars (for the milk category).
%For all datasets, the mean is greater than the median, which indicates right-skewed distributions. Meanwhile, the standard deviations and the maximum values are also very large, which indicates the long tails of the distributions. 
%%And the maximum values for most categories are more than 100 times greater than the mean. 
\lc{In Appendix 8 of the supplementary materials, we provide the summary statistics regarding the average sales amount of each single product (per store per week) for each dataset, as well as the weekly sales trends of the eight categories of products.}

%\begin{table}
%	%\begin{scriptsize}
%	\begin{footnotesize}
%		\caption{\footnotesize Descriptive statistics of the IRI weekly data (2008-2011) are displayed, which provides the minimum (Min), 1st quartile ($Q_1$), median, mean, 3rd quartile ($Q_3$), maximum (Max), and standard deviation (s.d.) of weekly sales for each product per store per week (in dollars).}
%		\begin{center}
%			
%			\begin{tabular}{cccccccc}
%				\hline
%				\hline
%				&Min		&$Q_1$	&Median	&Mean	&$Q_3$ &Max		&s.d.\\
%				\hline
%				%Beer			&0.01 & 12.90 & 24.57 & 52.69 & 50.97 & 22939.72 & 122.32	\\
%				Blades		&0.01 & 6.98 & 11.97 & 18.44 & 20.98 & 1611.96 & 23.08		\\
%				%Cigarettes		&0.01 & 8.80 & 21.56 & 46.95 & 47.95 & 42220.16 & 187.57		\\
%				Coffee	 	&0.00 & 7.58 & 13.58 & 25.04 & 26.97 & 12884.7 & 50.68	\\
%				Deodorant	 	&0.01 & 3.49 & 4.89 & 6.88 & 7.99 & 2227.80 & 7.94	\\
%				Diapers		&0.01 & 10.69 & 18.76 & 25.07 & 29.97 & 19345.97 & 31.63	\\
%				Frozen Pizza	&0.01 & 8.00 & 17.37 & 28.78 & 33.95 & 11375.20 & 43.79\\
%				Milk			&0.01 & 12.36 & 27.93 & 110.14 & 68.02 & 18379.83 & 353.05		\\
%				%Paper Towels	&0.01 & 15.97 & 34.93 & 71.84 & 72.45 & 66542.06 & 185.56\\
%				Photos		&0.01 & 7.49 & 11.97 & 17.42 & 20.97 & 742.6 & 18.60\\
%				Tooth Paste	&0.01 & 3.69 & 6.19 & 10.7 & 11.96 & 4013.1 & 19.89\\
%				\hline
%				\hline
%			\end{tabular}
%			
%		\end{center}
%		\label{summary}
%	\end{footnotesize}
%	
%\end{table}

%{\color{red}Revise national trend for 8 categories below}

\begin{figure}
	\centering \scalebox{0.345}{\includegraphics{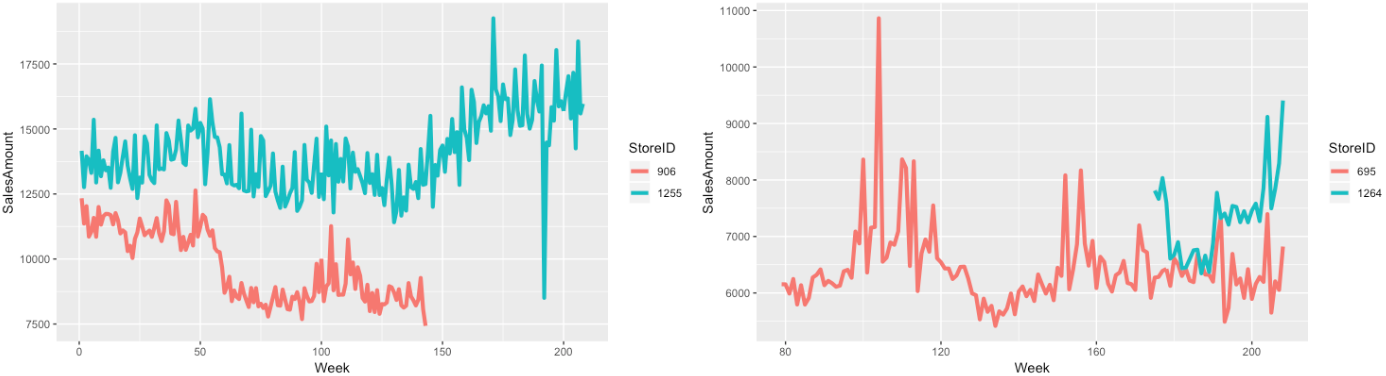}}
	\caption{\footnotesize Sales amounts of milk changed after the opening or closing of the sole competitor in the same region.}
	\label{compete}
\end{figure}

Since all the participating stores are grocery stores, we expect that local consumer demand dynamics exists among stores in close proximity. Therefore, we consider a store group as a geographical region\footnote{\scriptsize The same definition of ``region'' will be used in the rest of this article. A discussion of the advantages and disadvantages of this definition is provided in Section \ref{sec:summary}.} defined by the 3-digit zip code prefix.
%See Figure \ref{compete} for an illustration. 
As an illustration, in each panel of Figure \ref{compete}, we have identified a region in the Northeastern U.S. where only two chain stores were selling milk. In the left panel, store \#906 and \#1255 were the only two chain stores in the region. Store \#906 closed (or stopped selling milk) near the end of year 2010, and since then store \#1255 saw an increase in milk sales. In the right panel, store \#695 was once the only chain store in that region which sold milk. A new store, \#1264, opened (or started to sell milk) since spring 2011 whose sales amount started to increase by the end of summer 2011. Then we saw a slightly decrease in the milk sales in the existing store \#695. %This further shows the importance of incorporating local consumer demand (sales competition) information as part of forecasting methodologies.
\lc{However, product specific attributes (such as subcategory or volume) are not available in our datasets. Therefore, product demand dynamics is not considered in our numerical studies. In the Discussion section, we discuss this direction as an important future research area.}

\subsection{Model Description}
\lc{The proposed method is compared to seven methods from prior literature. One is the classical and widely-adopted time-series benchmark: the SARIMA model. We apply it to every store-product pair individually. In other words, a 208-week time series of sales amounts is analyzed for every single product in every single store
	%The SARIMA model is easy to implement, and does not require sales 
	without requiring data of other stores (or products). 
	%Hence we include it as our basic benchmark for comparison. 
	Another two methods are LSTM and vector autoregression \citep[VAR; e.g.,][]{holtz1988estimating} which can formulate multivariate time series.
	
	The other three methods are more recent and competitive latent factor models which have frameworks similar to the one of ATLAS and have demonstrated or claimed their capability of sales forecasting under the same or similar scenarios, i.e., where sales across multiple stores and products are available. Specifically, they are Bayesian probabilistic tensor factorization \citep[BPTF;][]{xiong2010temporal}, factorization machine \citep[libFM;][]{rendle2012factorization}, and temporal regularized matrix factorization \citep[TRMF;][]{yu2016temporal}. However, none of them consider local demand information in the latent factor estimation step. And while some of them consider extrapolation, none of them incorporate full SARIMA, or LSTM as canonical tools for forecasting. For VAR, LSTM, and TRMF, we apply them to each store individually, while for the rest of the latent factor models, we formulate all stores and products simultaneously.}

\iv{In addition, we also report the results of ATLAS without incorporating local demand dynamics, namely, CP decomposition (CPD) with SARIMA or LSTM, through which we demonstrate that incorporating local demand dynamics indeed improves the forecasting accuracy.} 
\lc{For the proposed method as well as most of the latent factor benchmarks, we expect sufficient observations from stores, products, or weeks (ideally close to or greater than the number of latent factors $k$ for each of them), such that the estimation procedures can be robust. This has been met in the IRI datasets.}

\subsection{Model Training and Validation}
%{\color{red}Mention an ANOVA is done before tensor decomposition. A sliding bar below to illustrate the proportion of training, tuning, and testing.}

\begin{figure}
	\centering \scalebox{0.63}{\includegraphics{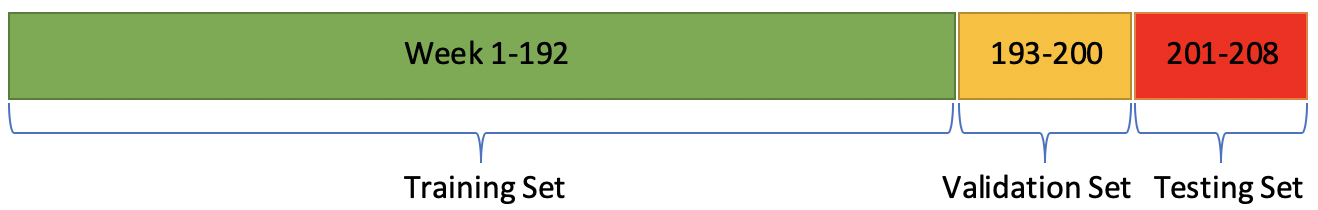}}
	\caption{\footnotesize Each dataset is split into a training set (week 1-192), a validation set (week 193-200) and a testing set (week 201-208), based on the chronological order.}
	\label{data_split}
\end{figure}

Recall that data from a total of 208 weeks are collected, and our ultimate goal is to forecast the future sales amounts based on historical sales data. Therefore, we split the training, validation, and testing set based on the chronological order, rather than a random split, to mimic the real-world forecasting scenario. Unlike in many applications where a series of rolling one-point-ahead forecasting is adopted \citep[e.g.;][]{osadchiy2013sales}, retail sales usually expect a longer-term forecasting for the sake of optimizing their operational decisions \citep{cui2018operational}. Therefore, we use data from the first 192 weeks as the training set, weeks 193-200 as the validation set, and the latest 8 weeks as the testing set (see Figure \ref{data_split} for an illustration). \lc{Summary of the training, validation, and test data, as well as detailed information about parameter tuning, its managerial interpretation, and model software availability, are provided in Appendices 9 and 10 of the supplementary materials.} Here the 8-week-ahead forecasting window was determined after some discussions with a few local experts, who are in retail business, subscribe to the IRI marketing data, and conduct sales forecasting on a regular basis. The length of forecasting horizon, however, can be adjusted based on different contexts or needs.
\lc{To evaluate the performance of ATLAS on different forecasting horizons, we allow a varying test data size (ranging from 4 to 20 weeks) in Appendix 12 of the supplementary materials, where ATLAS is compared with competing methods and demonstrates advantageous performance. We further conduct an additional experiment for robustness check in Appendix 13, where we evaluate ATLAS under different training data size (ranging from 8 to 192 weeks) to demonstrate its stability when the time range of training data collection is short.}

%\lc{(Provide specific tuning parameters here. Also mention different training and testing experiments here.)}

%Notice that, since the testing set has no temporal overlap with the training or the validation set, it is possible (and it indeed happened) that in the 8 weeks of the testing set new stores were opened, new products were released, and some (existing) stores started to sell products which they had never sold before, all of which are not available in the training or the validation set. This will be discussed in Section 6.

\section{Main Results: Product Sales Forecasting}

In this section, we apply ATLAS and competing methods to the eight IRI marketing datasets described in Section \ref{sec:data}.
%, and compare it with the four existing approaches discussed in Section 4.2. 
%The goal is to look at the general sales forecasting performance.
%, i.e., predicting the \xb{sales amount (in dollars)} of all products in each store for 8 future weeks.
%We demonstrate that latent factor models in general demonstrate high prediction accuracy. We also show that the proposed method, through incorporating sales competition, outperforms other competing latent factor approaches that were adapted to the product sales forecasting setting.

%First, we briefly demonstrate that the tensor based methods are more efficient than performing time series models individually. We aggregate store- and product-level data into region- and category-level, respectively. This guarantees that sales competition across regions or categories is light, and that sales given all region-category-week combinations are observed. Then applying CP decomposition with SARIMA is 3\% more accurate and 49\% faster than applying SARIMA individually.

\subsection{Performance Evaluation} \label{sec:performance}
For model performance evaluation, we utilize the popular and widely used numeric prediction accuracy metric -- root mean square error (RMSE), which assigns disproportionately larger penalties for bigger prediction errors.
%-- for numeric prediction to compare the proposed method with existing methods. 
Suppose $y_{ijt}$ is the sales amount of product $j$ in store $i$ at week $t$, and $\hat{y}_{ijt}$ is the predicted value of $y_{ijt}$. Then the RMSE on a given test set $\Omega$ is defined as
$\mbox{RMSE}(\Omega)=\left\{\frac{1}{|\Omega|}\sum_{(i,j,t)\in\Omega}(y_{ijt}-\hat{y}_{ijt})^2\right\}^{1/2},$
where $\Omega=\{(i,j,t): y_{ijt} \mbox{ is observed and }t>200\}$ in the IRI data context, and $|\Omega|$ is the size of $\Omega$.

%The key feature of the RMSE performance metric is that larger errors get disproportionately larger penalties (due to the square of the residual), thus, favoring models with a more consistent overall performance.

%Result tables interpretation
In Table \ref{results_RMSE}, we list the RMSE values exhibited by ATLAS and the competing methods.  Smaller RMSE indicates a smaller average distance between the predicted and the actual sales amounts. The differences in the results of CPD with SARIMA or LSTM are less than $10^{-4}$,
%\footnote{This is because the software package, Tensor Toolbox for Matlab, put small weights on the last mode (time-specific) of the tensor to avoid identifiability issues in their alternating least square algorithm.}, 
and hence are reported as one column. 

In general, ATLAS (either with SARIMA or LSTM) demonstrates the best performance across all product categories: ATLAS with SARIMA forecasting has the best performance in seven out of eight datasets, and ATLAS with LSTM forecasting has the best performance in the deodorant dataset.  It is possible that given a longer time series, LSTM would be able to achieve better performance. And in many categories, both ATLAS with SARIMA and ATLAS with LSTM perform better than the competing methods (or at least in the top 3).
\lc{It is important to note that, for VAR, the number of parameters is huge, which results in overfitting and sometimes unreasonably large predicted values for stores with few transaction records. To alleviate this issue and significantly improve the VAR predictive performance, we truncate the most extreme 10\% of VAR results, that is, replacing the largest (and smallest) 5\% predicted values by the 95th (and 5th) quantile.}
In particular, by comparing the results of CPD and ATLAS, we can see that incorporating local consumer demand dynamics indeed improves prediction accuracy. \lc{In Appendix 11 of the supplementary materials, we also consider performance evaluation under the mean absolute error metric, where ATLAS also demonstrates superior performance for the vast majority of datasets.} 
%\xb{(Space to discuss the performance of other methods?)}

\lc{To further evaluate the performance of ATLAS \emph{across} product categories, we combine ``coffee'' and ``milk'', as they both belong to beverages. We also combine ``blades'' and ``toothpaste'', as they both represent personal hygiene products. Furthermore, we combine ``coffee'', ``milk'', and ``frozen pizza'' as ``edibles'', and ``blades'', ``deodorant'', ``diapers'', ``photography'', and ``tooth paste'' as ``inedibles''. We demonstrate in Appendix 14 of the supplementary materials that, for products in categories which are more likely to be co-purchased, such as ``coffee'' and ``milk'', ATLAS is able to take advantage of the additional across-category information and further enhance its forecasting accuracy. In Appendix 15, we further compare ATLAS with competing methods after aggregating the original product-level data into completely category-level data (i.e., no more product-specific transactions), which significantly reduces data sparsity (i.e., much less data are ``missing'' at the store-category level). Advantageous performance of ATLAS is still observed, indicating its flexibility on datasets with varying degrees of density.}

%Specifically, in terms of RMSE in Table \ref{results_RMSE}, 
%\textcolor{red}{For categories like deodorant, diapers, and photography, some existing methods have performance very close to the proposed method.}
%Meanwhile, it can be seen that some traditional latent factor models report extremely high RMSEs. One possible reason is overfitting. 
%While the RMSE on the training set keeps reducing, the RMSE on the testing set increases drastically. 

%Additional results evaluated by the mean absolute error is provided in Table A2 in the Appendix.

\begin{table}
	\begin{footnotesize}
		\caption{RMSE results from the proposed method with SARIMA forecasting (ALTLAS(S)) and with LSTM forecasting (ATLAS(L)) are compared with baseline techniques, where VAR and LSTM are new benchmarks;	the lowest RMSE for each category is highlighted in bold.}
		\begin{center}
			\begin{tabular}{c|cccccc|c|cc}
				\hline
				\hline
				&SARIMA 	&VAR &LSTM &BPTF	&libFM	&TRMF	&CPD &ATLAS(S)	&ATLAS(L)\\
				\hline
				Blades		&17.09	&25.91 &18.53 &16.75	&21.03 	&17.65 	&17.17 &\textbf{14.26} 			&14.32\\
				Coffee		&60.47	&69.14 &56.08 &72.06	&54.83 	&61.97 	&59.03 &\textbf{52.32} 			&57.19\\
				Deodorant		&8.12	&7.96 &6.94 &9.48	&5.92	&7.23 	&5.57 &5.72 			&\textbf{5.54}\\
				Diapers		&30.01	&27.53 &26.10 &25.38	&16.34 	&25.48 	&17.31 &\textbf{15.67} 			&16.20\\
				Frozen Pizza	&41.46 	&47.37 &35.77&42.51 	&105.75	&43.00 	&39.33 &\textbf{33.92} 			&34.41\\
				Milk			&103.19 &336.44	&119.09	&204.31 	&349.70 	&108.85 &226.69 	&\textbf{78.91} 			&85.20\\
				Photography	&11.77 	&12.85&12.55&12.61 	&12.27 	&11.94 	&12.04&\textbf{10.66} 			&11.20\\
				Tooth Paste	&35.62	&21.00&18.96&48.25 	&28.38 	&22.98 	&18.18 &\textbf{17.17} 			&18.60\\
				\hline
				\hline
			\end{tabular}
			
		\end{center}
		\label{results_RMSE}
		
	\end{footnotesize}
\end{table}

\subsection{Efficiency}

%\textbf{In addition to performance, computational efficiency is another major concern especially when the sample size is large.} %Add this sentence in case reviewers want to compare with LR, ES and ARIMA for stores individually.
In terms of computational efficiency, all the latent factor models are significantly faster than the individual SARIMA, since the sales across multiple stores and products are modeled simultaneously.
Theoretically, BPTF, libFM, CPD, and ATLAS have the same computational complexity in terms of the number of stores, products, and weeks, as all of them estimate the three modes of a tensor iteratively and their loss functions have similar forms with explicit solutions. TRMF, VAR, and LSTM are applied to each store individually, and are hence slower than the tensor based methods. In general, the computational complexity of each method is provided in Table \ref{complexity}. When the number of stores and products are large, BPTF, libFM, CPD, and ATLAS are expected to have high computational efficiency.

\begin{table}
	\begin{footnotesize}
		\caption{\footnotesize Computational complexity of the tensor decomposition of ALTAS is compared with baseline techniques. Here $n_s$, $n_p$ and $n_w$ represent the number of stores, products, and weeks, respectively.}
		\begin{center}
			\begin{tabular}{ccc}
				\hline
				\hline
				ARIMA		&TRMF, VAR and LSTM &ATLAS and others\\
				\hline
				$O(n_sn_pn_w)$	&$O(n_s(n_p+n_w))$ &$O(n_s+n_p+n_w)$\\
				\hline
				\hline
			\end{tabular}
			
		\end{center}
		\label{complexity}
		
	\end{footnotesize}
	
\end{table}

%\nd{provide complexity O() table here.}

%\nd{Provide comp time for toothpaste and fzpizza as examples. mention overfitting.}

In practice, however, the computational runtime (model building and deployment) may differ, as the best-performing model parameters, convergence criterion, and number of iterations might be pre-specified differently. For example, for the frozen pizza dataset, the computational runtimes for individual SARIMA, VAR, LSTM, BPTF, libFM, TRMF, CPD, and ATLAS were 28.07, 9.13, 1.62, 0.80, 0.93, 5.91, 0.42, and 1.83 hours, respectively. And for the toothpaste dataset, the computational runtimes for individual SARIMA, VAR, LSTM, BPTF, libFM, TRMF, CPD, and ATLAS were 18.49, 10.90, 8.55, 0.42, 0.79, 4.51, 0.23, and 0.81 hours, respectively. It can be seen that the computational runtimes for each method roughly follow the theoretical complexity as provided in Table \ref{complexity}, where SARIMA is the longest, followed by VAR, LSTM, and TRMF, while BPTF, libFM, CPD, and ATLAS are at comparable levels. 
%And in particular, ATLAS is slower than CPD due to the incorporated local consumer demand.

\section{Summary and Conclusions} \label{sec:summary}

%\nd{Advantages of the proposed method: incorporates sales information about other stores and other products, and formulates sales competition in the model, both of which improves forecasting accuracy.} 

%\subsection{\xb{Summary and Managerial Implications}}
Following the design science paradigm, we propose, develop, and evaluate a new latent-factor-based approach, namely the Advanced Temporal Latent-factor Approach to Sales forecasting (ATLAS). Specifically, ATLAS is designed for the complex settings where sales data from multiple stores across multiple products are collected, and sales forecasting for many store-product combinations is of interest. The key feature of our work is that ATLAS is able to incorporate elements of local consumer demand information in a large-scale, multiple-store, multiple-product setting. It leads to significant accuracy improvements, especially compared with existing latent factor models which use a similar framework.
Such accuracy improvement on sales forecasting provides important managerial implications for companies' decision making for budgeting, production, supply chain management, and inventory control. 

This work opens up several interesting opportunities for future studies. 
\lc{Although the proposed model 
	%is purely a data-driven, machine learning model, which 
	improves upon existing methods through incorporation of local demand dynamics as part of its regularization procedure,
	%. However, as discussed in Section \ref{sec:demand}, 
	it does not attempt to model the consumer demand directly, e.g., based on sophisticated economic theory.} While the current approach does provide substantial forecasting accuracy improvements, we believe that incorporating economic theory considerations into machine learning models can provide significant additional advantages, and, thus, constitutes a promising direction for future work. For instance, from a microeconomic perspective, the consumer demand might be formulated by the consumer choice models as in demand theory \citep{bohm2008demand}. Collecting data at the customer level and incorporating consumer choice models into the tensor framework might lead to further improvements in forecasting accuracy.

\lc{Second, refining the definition of the demand dynamics may further improve flexibility and forecasting accuracy of the proposed model. For example, there could be asymmetric competition among stores or products, which may not be formulated as correlation matrices. Therefore, new regularization methods to accommodate asymmetric matrices are needed. Furthermore, a time-dependent demand formulation may be considered. This requires the estimation of time-varying parameters as well as imposes additional assumptions regarding how the covariance matrix would change in the near future. To address this issue, one possible solution is to allow the demand dynamics to change less rapidly, for example, on a seasonal or yearly basis, based on domain knowledge, such that fewer new parameters are needed while the demand pattern is no longer static.
	%consumers living at the boundary of two regions may purchase products from grocery stores at either side of the boundary, and hence consumer demand across adjacent regions is not considered. 
	Meanwhile, we are also aware that there may be other store competitors, both online and offline, which can be considered if their data become available.
	
	%Second, collecting and analyzing important store- and product-specific attributes 

	Finally, 
	%aside from the data-driven approaches to forecasting that are based directly on sales data, 
	there can be additional socioeconomic, marketing, managerial, operational, and product-related attributes which contribute to the forecasting results. For example, the IRI dataset provides demographic information of the population within 2-mile radius of each store location, which includes, for example, number of households, average age, percentage of males/females, and average household income. Incorporating this information may further explain demand dynamics and enhance sales forecasting. Furthermore, it would be interesting to explore which specific products (or what specific characteristics of products) favor ATLAS. Collecting product-related information may help address this issue and provide deeper managerial implications regarding where and how the proposed method may be most advantageous. Meanwhile, if other variables, such as managerial and operational strategies are available, one may consider generalizing the proposed approach even further to utilize this information.}

\section*{Acknowledgement}
The authors thank Mike Kruger and the Information Resource, Inc (``IRI'') for providing the IRI marketing data. The research is partially supported by NSF Grants DMS-1613190 and DMS-1821198.

%\clearpage
\bibliographystyle{myapalike}
\bibliography{ATLAS_arxiv}

%%%%%%%%%%%%%%%
\end{document}